\documentclass{article} % For LaTeX2e
\usepackage{iclr2027_conference,times}

\usepackage{amsmath,amsfonts,bm}

\def\eqref#1{equation~\ref{#1}}
\def\1{\bm{1}}

\DeclareMathAlphabet{\mathsfit}{\encodingdefault}{\sfdefault}{m}{sl}
\SetMathAlphabet{\mathsfit}{bold}{\encodingdefault}{\sfdefault}{bx}{n}

\usepackage{hyperref}
\usepackage{cleveref}
\usepackage{url}
\usepackage{lipsum}
\usepackage{colortbl} % \rowcolor
\definecolor{ourgray}{gray}{0.93}
\usepackage{tabularx}
\usepackage{graphicx}
\usepackage{wrapfig}
\usepackage{enumitem}
\usepackage{subcaption}
\usepackage{float}

\title{Cache locally, recompute selectively\\for streaming video understanding}

\author{Junhyeok Lee, Sangyoon Lee, Jaeho Lee \\
Pohang University of Science and Technology (POSTECH) \\
\texttt{\{jhlee0310,sangyoon.lee,jaeho.lee\}@postech.ac.kr} \\
}

\usepackage{xcolor}
\definecolor{linkblue}{RGB}{0,105,180}
\hypersetup{
    colorlinks=true,
    allcolors=linkblue,
    allbordercolors={0 0.4118 0.7059},
    pdfborder={0 0 0}
}

\iclrfinalcopy % Uncomment for camera-ready version, but NOT for submission.

\begin{document}

\maketitle

\begin{abstract}

Streaming video understanding (SVU) requires video-LLMs to process incoming video continuously while answering user queries with low latency. Recent methods reduce query-time cost by prefilling frames into KV caches during idle time and reusing the cached KVs at query time. To bound per-frame computation over unbounded video streams, these methods restrict the context available during prefill. This creates a fundamental context mismatch: the restricted context may omit information needed to properly contextualize video tokens, yet, still contain information irrelevant to the eventual query, which is unknown during prefill. We propose \textbf{FrameBlend}, a simple yet effective plug-and-play method to address this mismatch through selective KV recomputation, without requiring any additional training. At query time, FrameBlend re-prefills selected video tokens using query-relevant context and replaces their cached KVs with recomputed ones. Crucially, this recomputation enables shorter attention windows during idle-time prefill, substantially reducing prefill cost while improving accuracy. Experiments across three SVU frameworks and five streaming and offline benchmarks demonstrate accuracy gains of up to 3.9 pp. On RVS-Ego, FrameBlend reduces per-frame prefill FLOPs by over 20\% and total processing time by 14--59\% for retrieval-based frameworks, including the cost of query-time recomputation. These results demonstrate that a simple change in how contextualization is distributed between idle time and query time can improve the accuracy-efficiency trade-off for SVU.

\end{abstract}

\section{Introduction} \label{sec:introduction}

Recent video large language models (video-LLMs) have demonstrated strong video understanding capabilities \citep{li2024llava,bai2025qwen3,bai2025qwen25vltechnicalreport}. These advances, together with growing demand for real-time applications such as robotics \citep{mu2023embodiedgpt}, autonomous driving \citep{shao2024lmdrive}, and surveillance systems~\citep{rao2022real}, have spurred interest in streaming video understanding (SVU). In SVU, a video-LLM continuously processes an incoming video stream and answers user queries as they arrive. This setting imposes three constraints. First, during \textit{idle time}, i.e., no query is pending, the model must keep pace with a potentially unbounded stream under limited compute and memory. Second, because future queries are unknown during idle-time processing, it is difficult to decide in advance which information to preserve. Third, at \textit{query time}, the model must respond with minimal latency, leaving little room for additional computation before answering.

To meet these requirements, recent methods prefill incoming frames into key-value (KV) caches during idle time and directly reuse the cached KVs at query time \citep{di2025streaming,ning2025livevlm,zhang2026hermes}. However, letting every incoming token attend to all preceding tokens incurs a cumulative attention cost that grows quadratically with stream length, which is prohibitive for long streams. Existing methods therefore limit the prefill context, either through sliding-window attention, which restricts attention to recent frames \citep{di2025streaming,chen2026streamingtom,chen2026streamkv}, or through KV cache compression, which condenses accumulated context into a fixed-size memory~\citep{kim2026infinipot,ning2025livevlm,zhang2026hermes}.

Both strategies can create a \textit{mismatch} between the context used to construct KVs during idle-time prefill and the context used at query time. With sliding-window attention, a token cannot attend to earlier tokens outside its window, even if those tokens are later retrieved together with it to answer a query. With KV cache compression, information from tokens that are later compressed or evicted may remain encoded in retained KVs, even when it is irrelevant to the query \citep{pang2026decouple}. Reused KVs may thus both omit relevant context and retain irrelevant context. Direct KV reuse leaves these representations uncorrected, potentially degrading answer quality  \citep{yao2025cacheblend}.

\begin{figure}[t]
\vspace{-0.5em}
\begin{center}
\includegraphics[width=0.98\linewidth]{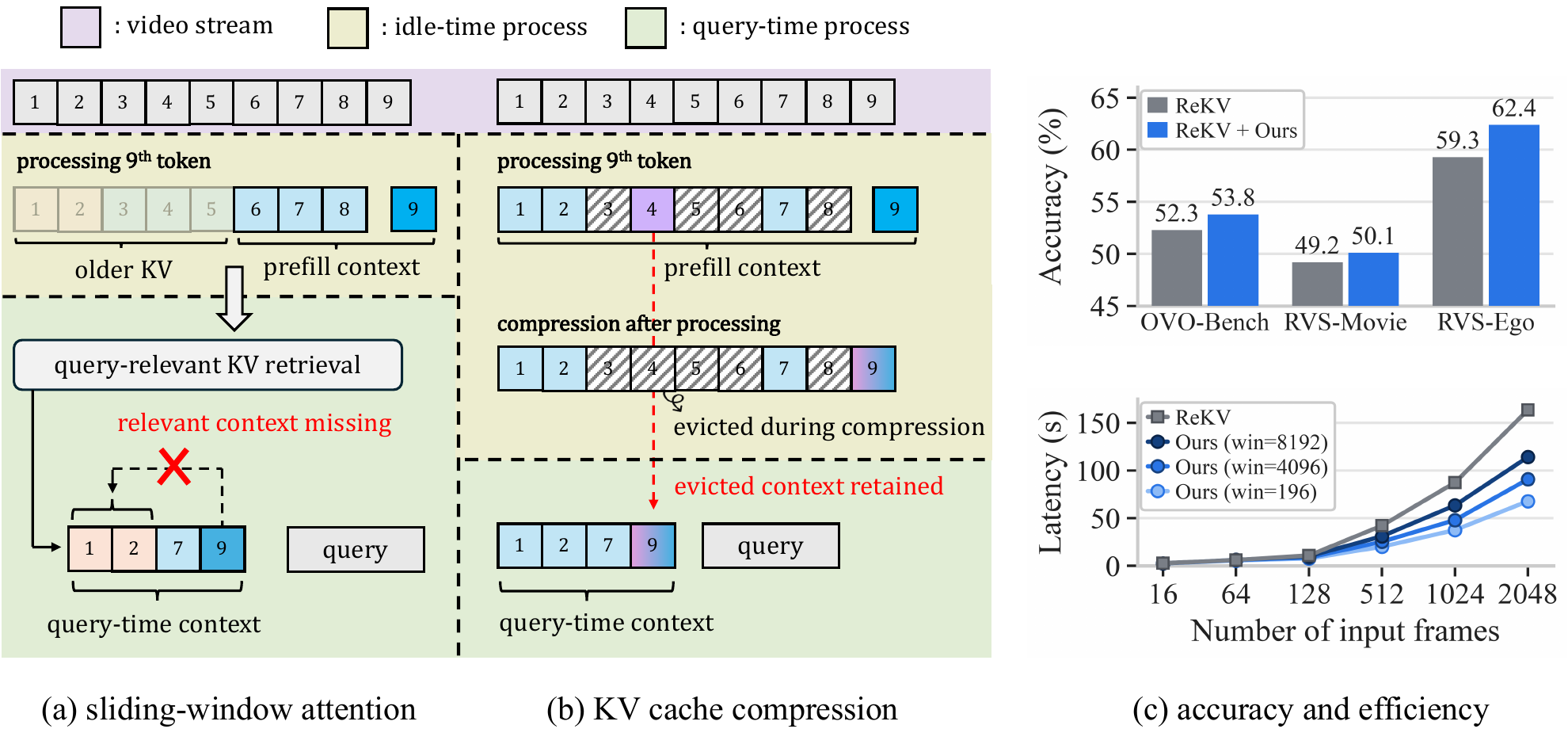}
\end{center}
\vspace{-1em}
\caption{\textbf{Motivation and performance of FrameBlend.}
(a) Sliding-window attention can omit inter-token context needed at query time.
(b) KV compression can retain unnecessary context from evicted tokens.
(c) FrameBlend improves SVU framework accuracy while reducing total latency.}
\label{fig:motivation}
\end{figure}

To resolve this mismatch, we propose \textbf{FrameBlend}, a training-free method that selectively recomputes cached KVs at query time. Inspired by recomputation techniques in LLM serving \citep{yao2025cacheblend,hu2024epic}, FrameBlend re-prefills a selected subset of tokens under the query-time context and overwrites their cached KVs. Whereas prior methods either rely on costly operations to identify recomputation targets or select them in a query-agnostic manner, FrameBlend uses QTS (query-relevant token-level selection), a lightweight, frequency-based mechanism guided by query relevance. Moreover, in SVU, query-time recomputation fundamentally changes how idle-time prefill should be designed; rather than using large attention windows to preserve context in advance, FrameBlend instead uses shorter windows, as missing context can be recovered once the query arrives. This not only reduces prefill cost but can also improve accuracy over longer windows. We attribute this gain to reduced contextual interference: since future queries are unknown at prefill time, shorter windows limit how much query-irrelevant information is baked into the cached KVs. Together, \textit{shorter attention windows} during idle-time prefill and \textit{query-time KV recomputation} form the core principle of FrameBlend, improving both efficiency and accuracy without training, and with minimal modifications to existing SVU frameworks.

We evaluate FrameBlend on three SVU frameworks---ReKV~\citep{di2025streaming}, HERMES~\citep{zhang2026hermes}, and LiveVLM~\citep{ning2025livevlm}---across two streaming benchmarks: RVS~\citep{zhang2024flash} and OVO-Bench~\citep{niu2025ovo}. We also assess general video understanding ability on three offline benchmarks \citep{fu2025video,zhou2025mlvu,mangalam2023egoschema}. FrameBlend matches or improves aggregate accuracy on every framework-benchmark pair, with gains of up to 3.2 pp on streaming benchmarks and 3.9 pp on offline benchmarks. For ReKV and LiveVLM, FrameBlend further reduces per-frame prefill FLOPs by 22.2--26.0\% and prefill latency by 12.6--60.5\% on RVS. Even after accounting for query-time recomputation, total FLOPs and latency decrease by 19.0--23.2\% and 14.1--58.7\%, respectively.

Our contributions are summarized as follows:
\begin{itemize}[leftmargin=*,topsep=0pt,parsep=0pt]
    \item We propose FrameBlend, a plug-and-play, training-free method that selectively recomputes KVs for relevant tokens at query time. 
    To our knowledge, it is the first query-aware recomputation method for SVU.
    % To our knowledge, it is the first recomputation method for SVU frameworks that separate inference into idle-time prefill and query-time decoding.
    \item We show that query-time recomputation enables shorter attention windows during idle-time prefill, reducing prefill cost while matching or exceeding the accuracy of longer windows.
    \item Across ReKV, HERMES, and LiveVLM on two streaming and three offline benchmarks, FrameBlend consistently improves (or matches) aggregate accuracy for every framework--benchmark pair, with gains of up to 3.2 pp and 3.9 pp, respectively. For ReKV and LiveVLM on RVS, it also reduces per-frame prefill FLOPs by 22.2--26.0\% and total latency by 14.1--58.7\%.
\end{itemize}

\section{Related work} \label{sec:related_work}

\paragraph{Streaming video understanding.} The central challenge in SVU is efficiently storing and utilizing the information in potentially unbounded video streams, under memory and computation constraints. One line of work addresses this by maintaining an explicit memory of visual representations from incoming frames~\citep{zhang2024flash,qian2025dispider,wang2026streambridge,zeng2026streamforest}. However, these stored representations must still be processed by the backbone LLM to answer the incoming query, often leading to long query-time latency for LLM prefill computation.

Recently, a growing line of work instead prefills incoming frames into KV cache during idle time and directly reuses the cached KVs at query time, thereby shifting much of the prefill computation to idle time~\citep{di2025streaming,chen2026streamkv,chen2026streamingtom,kim2026infinipot,yang2025streammem,ning2025livevlm,zhang2026hermes}.
These approaches can be categorized according to how they manage the context during the prefill and decoding.
During prefill, allowing each frame to attend to all preceding frames becomes prohibitively expensive as the stream grows.
% Existing methods therefore adopt various strategies to keep the growing context within bounded budgets.
Therefore, some approaches restrict attention to recent frames using sliding-window attention~\citep{di2025streaming,chen2026streamkv,chen2026streamingtom}, while others apply KV cache compression to reduce redundant tokens~\citep{kim2026infinipot,yang2025streammem,ning2025livevlm,zhang2026hermes}.
During decoding, retrieval-based methods independently select query-relevant KVs at each layer based on their similarity to the query~\citep{di2025streaming,chen2026streamingtom,ning2025livevlm}.
In contrast, non-retrieval methods use the entire bounded KV memory, which has been progressively compressed during prefill, to generate the final prediction at the query time~\citep{kim2026infinipot,yang2025streammem,zhang2026hermes}.

These works directly reuse the stored KVs for decoding, which can create a context mismatch between prefill and decoding. Our work mitigates this mismatch by selectively recomputing cached KVs with query-relevant context at query time.

\paragraph{Selective recomputation of KV cache.} The idea of recomputing the cache for resolving context mismatch has been studied recently in the field of long-context LLMs. In these scenarios, the primary goal is to approximate the behavior of full-context KVs, by recomputing only a fraction of the KVs that have been previously prefilled by computing in short chunks. For example, CacheBlend dynamically selects and recomputes the KVs for tokens that deviate large by such recomputations \citep{yao2025cacheblend}.
EPIC instead statically recomputes the first few tokens of each cached chunk~\citep{hu2024epic}. These ideas have been extended to vision-language model serving, where MPIC recomputes the first few image tokens~\citep{zhao2025mpic}, while VLCache dynamically adjusts the recomputation ratio across layers~\citep{qin2025vlcache}.

These methods, however, are not specifically designed for SVU, where recomputation must be both query-aware and efficient under a tight query-time latency budget.
Their recomputation targets are determined either by query-agnostic heuristics or by additional query-time operations. In contrast, FrameBlend selects query-relevant recomputation targets and leverages existing backbone retrieval results where applicable, thereby minimizing additional target-selection operations.

\section{Preliminaries} 
\label{sec:preliminaries}

In this section, we formally describe the problem of streaming video understanding and the idea of KV cache reuse. For the sake of simple exposition, we focus on describing the case of ReKV---a method with sliding-window prefill and decoding based on selective retrieval \citep{di2025streaming}.

Consider a standard streaming video question-answering task, where the model is given a stream of video frames. More concretely, define the \textit{video token stream} as an unlimited sequence of tokens
\begin{align}
\mathcal{X}=(\mathbf{x}_1,\mathbf{x}_2,\mathbf{x}_3,\ldots),
\end{align}
where the tokens are temporally ordered. That is, $\mathcal{X}$ consists of the tokenized version of each video frame, concatenated in the temporal order. The \textit{user query} $\mathbf{q}$  may arrive in the middle of this stream, e.g., after the first $N$ video tokens, asking a specific question about the video. Then, the video-LLM must generate an answer $\hat{y}$ using the observed token sequence $\mathcal{X}_{\le N} = (\mathbf{x}_1,\ldots,\mathbf{x}_N)$.

Because storing all tokens $\mathcal{X}_{\le N}$ and processing them after the query arrives incurs too much latency and storage requirements, we consider KV cache reuse \citep{di2025streaming,ning2025livevlm,zhang2026hermes}. Concretely, we divide the streaming inference into two phases: \textit{idle time} and \textit{query time}, corresponding to before and after the query arrives, respectively.

\textbf{Idle time.} During idle time, each incoming video token $\mathbf{x}_n$ is processed continually, forming the (layerwise) \textit{memory} $\mathcal{M}_{n}^{\ell}$ for each transformer layer $\ell \in \{1,\ldots,L\}$. For standard sliding-window methods, e.g., ReKV \citep{di2025streaming}, this is simply done by prefilling $w$ most recent tokens $(\mathbf{x}_{n-w+1},\ldots,\mathbf{x}_{n})$ to the video-LLM and storing the key-value (KV) cache $\mathbf{c}^{\ell}_n = (\mathbf{k}_n^{\ell},\mathbf{v}_n^{\ell})$ from each layer. That is, the memory will keep accumulating these KV caches as
\begin{align}
    \mathcal{M}_{n}^{\ell} = (\mathbf{c}^{\ell}_1,\mathbf{c}^{\ell}_2,\ldots,\mathbf{c}^{\ell}_n).
\end{align}
More recent methods employ mechanisms to continually update and compress the previously generated memory, keeping the total storage cost sublinear in $n$ \citep{ning2025livevlm,zhang2026hermes}.

\textbf{Query time.} When the query $\mathbf{q}$ arrives at time $N$, the model utilizes the current memory $\{\mathcal{M}_{N}^{\ell}\}_{\ell}$ to generate the response, rather than prefilling the entire observed sequence from scratch. To reduce the cost of loading the memory, one can first select a subset of KV caches $\mathcal{S}^{\ell} \subseteq \mathcal{M}_{N}^{\ell}$ for each layer $\ell$, that is most relevant for answering the target query $\mathbf{q}$. This is typically done by applying some similarity metric $\rho(\cdot,\cdot)$ between the query and the keys of each KV, i.e.,
\begin{align}
\mathcal{S}^{\ell} = \left\{(\mathbf{k}^{\ell}_i,\mathbf{v}^{\ell}_i)~\big|~\rho(\mathbf{k}^{\ell}_i,\mathbf{q}) \in \mathrm{ArgTopK}\big(\{\rho(\mathbf{k}^{\ell}_i,\mathbf{q})\}_{i=1}^N\big)\right\}. \label{eq:query_time_similarity}
\end{align}
For example, many prior works use the dot-product similarity between the stored key and the query representation derived from $\mathbf{q}$ \citep{di2025streaming,chen2026streamingtom,ning2025livevlm}. In contrast, the methods that compress-and-store the KV during idle time typically utilize the whole memory---i.e., $\mathcal{S}^{\ell} = \mathcal{M}_{N}^{\ell}$---as it is already compact \citep{kim2026infinipot,yang2025streammem,zhang2026hermes}. Critically, we note that these KV subsets $\mathcal{S}^{\ell}$ do not need to correspond to an identical set of tokens across different layers; in fact, different layers commonly select different tokens.

\section{Method} \label{sec:method}

\begin{figure}[t]
\begin{center}
\includegraphics[width=0.95\linewidth]{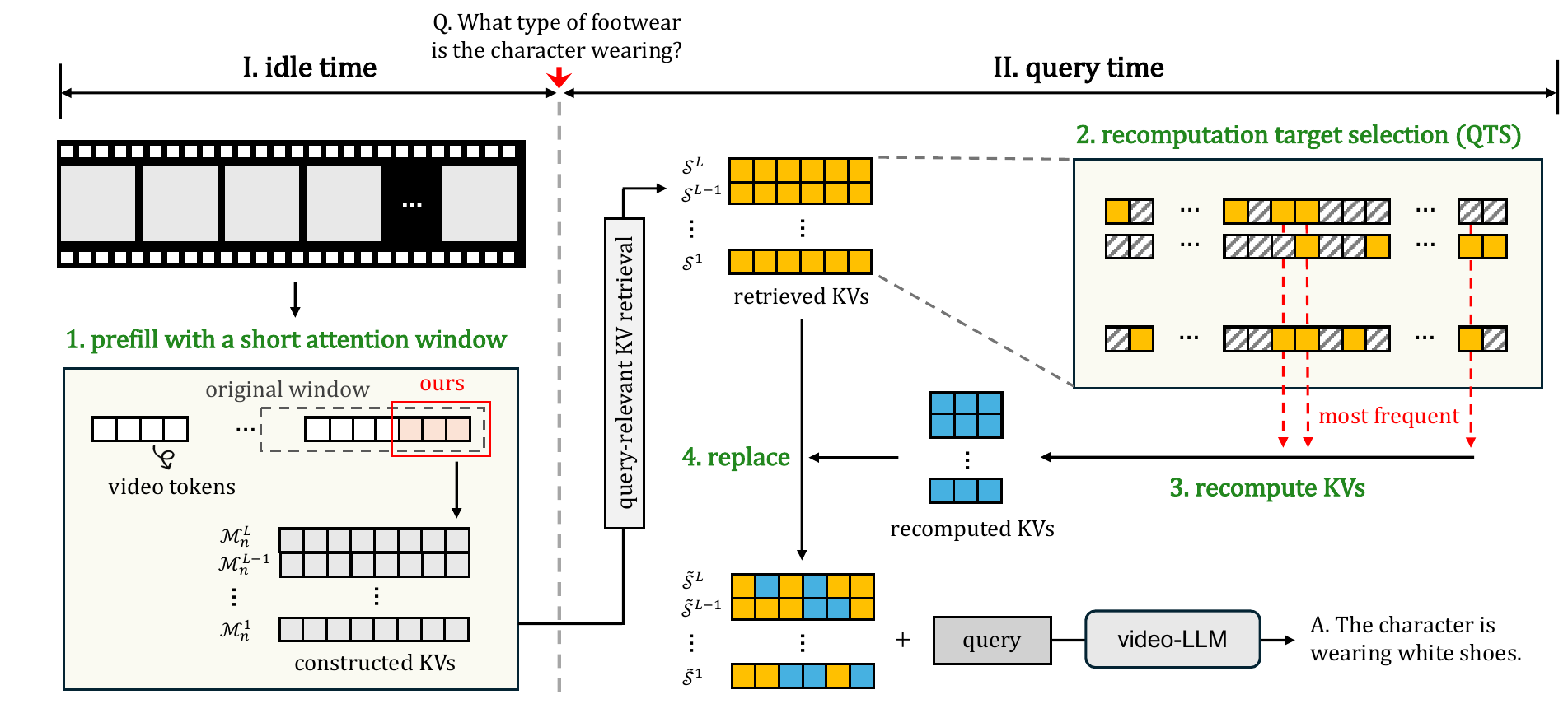}
\end{center}
\caption{
\textbf{Overview of FrameBlend.}
FrameBlend uses a short attention window during idle-time prefill and selectively recomputes the KVs of query-relevant tokens at query time.
}
\label{fig:method}
\end{figure}

We propose \textbf{FrameBlend}, a simple yet effective plug-and-play method to reduce the prefill cost and improve the accuracy of SVU frameworks by selectively recomputing the KV cache at query time. We first formally describe how FrameBlend works (\Cref{ssec:recompute}), and discuss how it allows us to use a shorter attention window (\Cref{ssec:short_window}). By default, we focus on describing the case of applying on retrieval-based methods, e.g., ReKV \citep{di2025streaming} or LiveVLM \citep{ning2025livevlm}. Then, we additionally discuss how to adapt it for KV compression methods, e.g., \citet{zhang2026hermes}.

\subsection{Selective Recomputation of KV Cache} \label{ssec:recompute}

After the base SVU framework (e.g., ReKV) determines the layerwise KV sets $\{\mathcal{S}^\ell\}_{\ell=1}^{L}$ for answering the query $\mathbf{q}$, FrameBlend operates in two simple steps: (1) \textit{Recomputation target selection:} We select a subset of the loaded KV sets to recompute; (2) \textit{Recomputing and decoding:} We recompute the KVs and use them for decoding the answer. We describe each step below.

\paragraph{Step 1: Recomputation target selection.} In this step, we select a subset of the retrieved KV cache $\mathcal{S}^{\ell}$ that will be recomputed, using the QTS (query-relevant token-level selection) criterion.

The design principle of QTS is twofold. First, we prioritize the most query-relevant KVs, because they will have the largest impacts on the final answer. Second, we select KVs in a way that recomputing can be done efficiently, in order to minimize the query-time latency. Typical recomputing methods for LLM serving, e.g., CacheBlend \citep{yao2025cacheblend}, are suboptimal in this sense; they focus on approximating the properties of full KV cache by recomputing KVs that get updated the most, gradually proceeding over the layers. This is not aligned well for SVU, being query-agnostic and 
requiring a tedious sequential selection procedure between transformer layers.

In short, QTS makes selection at the token-level, i.e., select KVs corresponding to the same tokens at every layer. This design is geared toward minimizing the latency, by enabling an allocation of constant computational budget over the layers and removing any inter-layer operations.

% With this token-level selection, since recomputing a token at layer $\ell$ requires forwarding it through all preceding LLM layers, the effect of recomputing a token is largely independent of how many layers retrieve it.
% For example, recomputing a token selected by only a few layers incurs the same full forward-pass cost while updating KVs in only a small fraction of the entire retrieved tokens.
% To maximize the coverage of recomputation under a fixed computational budget, QTS prioritizes tokens that are selected most frequently across layers.

QTS selects the tokens that have been most frequently selected as $\mathcal{S}^{\ell}$ across all layers. More concretely, the set of \textit{recomputation target tokens} is defined as
\begin{align}
\mathcal{P} = \left\{\mathbf{x}_i~\big|~\phi(\mathbf{x}_i) \in \mathrm{ArgTopK}\big(\{\phi(\mathbf{x}_i)\}_{i=1}^N\big)\right\},\qquad \mathrm{where}\quad
\phi(\mathbf{x}_i)
:=
\left|
\left\{
\ell :
\mathbf{c}_i^\ell \in \mathcal{S}^\ell
\right\}
\right|.
\end{align}
Since this $\mathcal{P}$ is computed directly from the backbone framework's selections, it captures the cross-layer agreement on the query relevance of $\mathbf{x}_i$ without additional operations.

For frameworks without query-time retrieval, e.g., HERMES~\citep{zhang2026hermes}, QTS simply applies the same similarity-based retrieval to the tokens in memory (\Cref{eq:query_time_similarity}), and chooses the highest-similarity tokens. The results of this retrieval only affect which tokens to recompute, and the backbone's original decoding context remains unchanged.

\paragraph{Step 2: Recomputing and decoding.}

The tokens in $\mathcal{P}$ are jointly re-prefilled through all LLM layers in temporal order, which requires a single forward pass over $|\mathcal{P}|$ video tokens.
The recomputed KVs then replace the corresponding KVs in $\{\mathcal{S}^{\ell}\}_{\ell=1}^L$ to construct the final decoding context. Concretely, let $\hat{\mathbf{c}}_{i}^{\ell}$ denote the recomputed KV of token $i$ at layer $\ell$.
The replacement is defined as
\begin{equation}
\tilde{\mathbf{c}}_i^\ell =
\begin{cases}
\hat{\mathbf{c}}_i^\ell, & \mathbf{x}_i \in \mathcal{P},\\
\mathbf{c}_i^\ell, & \text{otherwise},
\end{cases}
\qquad
\tilde{\mathcal{S}}^\ell
=
\left\{
\tilde{\mathbf{c}}_i^\ell
\mid
\mathbf{c}_i^\ell \in \mathcal{S}^\ell
\right\}.
\end{equation}
The answer is decoded using $\tilde{\mathcal{S}}^\ell$, following the same temporal ordering and positional encoding policy as the backbone SVU framework.
Additionally, to avoid repeated vision-encoder computation during recomputation, FrameBlend retains each observed video token in CPU memory during idle-time prefill.

\subsection{Prefilling with a short attention window} \label{ssec:short_window}

\begin{wrapfigure}[11]{r}{0.45\textwidth}
\vspace{-\intextsep}
\begin{centering}
%\framebox[4.0in]{$\;$}
\includegraphics[width=\linewidth]{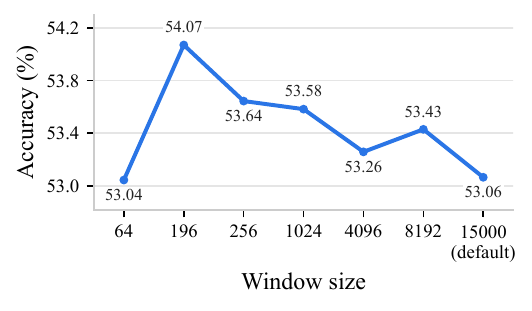}
\end{centering}
\vspace{-2em}
% \caption{Accuracy of ReKV+FrameBlend on OVO-Bench, for varying window sizes.} 
\caption{\textbf{Effect of window size.} Accuracy of ReKV with FrameBlend on OVO-Bench.}
\label{fig:window_size_ablation_ovo}
\end{wrapfigure}

With post-query recomputation, we revisit the amount of context required during prefill.
Since recomputing provides inter-token context at the query time, FrameBlend may need less contextual information during the idle time. In other words, we may need a shorter window for the prefill.

Our experimental results confirm this intuition. In \Cref{fig:window_size_ablation_ovo}, we report the accuracy of FrameBlend applied on ReKV with various window sizes, measured on the OVO-Bench for LLaVA-OV-7B. For simplicity, we have used $K=\infty$ for recomputing (i.e., recompute all retrieved cache).

From \Cref{fig:window_size_ablation_ovo}, we observe that reducing prefill windows can improve accuracy for FrameBlend, as short as 196 tokens---which correspond to a single video frame. This is in stark contrast with the observations of ReKV, where a longer window generally leads to better performance \citep{di2025streaming}. We attribute this gain to reduced contextual interference at prefill~\citep{pang2026decouple,shen2026simple}: since the future query is unknown, a longer context increases the risk of encoding unnecessary contextual information into the KVs. 
Results on additional benchmarks are provided in Appendix~\ref{app:window_size_sweep}.

FrameBlend thus reduces the attention window to 1--68\% of the backbone's original window size, depending on the benchmark,
reducing prefill compute and improving accuracy over longer ones.

\section{Experiments} \label{sec:experiment}
\subsection{Experimental setup} \label{ssec:experimental_setup}

\textbf{Benchmarks.}
We evaluate FrameBlend on five streaming and offline video understanding benchmarks, covering both multiple-choice and open-ended question answering.
For streaming evaluation, we use OVO-Bench~\citep{niu2025ovo} and RVS~\citep{zhang2024flash}. On RVS, following \citet{zhang2026hermes}, we use GPT-3.5-turbo-0125 to evaluate open-ended answers.
For offline evaluation, we use MLVU~\citep{zhou2025mlvu}, Video-MME~\citep{fu2025video}, and EgoSchema~\citep{mangalam2023egoschema}.
We use the official development set for MLVU, the full EgoSchema test set evaluated by the official evaluation server, and Video-MME without subtitles.

\textbf{Base SVU methods.} To evaluate FrameBlend as a plug-and-play method, we integrate it into three frameworks: ReKV~\citep{di2025streaming}; LiveVLM~\citep{ning2025livevlm}; HERMES~\citep{zhang2026hermes}. We retain the original configuration of each framework except for applying the query-time recomputation and the short prefill window introduced by FrameBlend.

\textbf{Video-LLMs.} We evaluate on two open-weight backbone video-LLMs: LLaVA-OneVision-7B (LLaVA-OV-7B)~\citep{li2024llava} and Qwen2.5-VL-7B~\citep{bai2025qwen25vltechnicalreport}.

\textbf{Hyperparameters.} All video inputs have been sampled at 0.5 FPS.
We set the recomputation budget to $K=7840$ video tokens, corresponding to 40 frames for LLaVA-OV-7B. For FrameBlend, we have tuned prefill window sizes for each framework-benchmark pair.

\textbf{Hardware.} All efficiency experiments have been conducted on a single NVIDIA A100 80GB GPU.

\textbf{Omitted details.} See Appendix~\ref{app:implementation_details}.

\subsection{Main results} \label{ssec:main_result}

\begin{table}[t]
\caption{\textbf{FrameBlend on streaming benchmarks.} We compare the accuracy of various SVU frameworks with and without FrameBlend on streaming VQA benchmarks.}
\label{tab:acc_rvs_ovo_a5000_opt}
\begin{center}
\small
\renewcommand{\arraystretch}{1.30}
\setlength{\tabcolsep}{6.25pt} 
\begin{tabular}{l c cc c cc c ccc} 
\noalign{\hrule height 0.6pt}
 & & \multicolumn{2}{c}{\bf RVS-Ego} & & \multicolumn{2}{c}{\bf RVS-Movie} & & \multicolumn{3}{c}{\bf OVO-Bench} \\
\cline{3-4} \cline{6-7} \cline{9-11}
{\bf Method} & & Acc. & Score & & Acc. & Score & & Real-time & Backward & Avg. \\
\hline
\noalign{\renewcommand{\arraystretch}{1.0}}
\multicolumn{11}{l}{\bf LLaVA-OV-7B} \\
\noalign{\renewcommand{\arraystretch}{1.5}}
\hline
ReKV & & 59.3 & 3.93 & & 49.2 & \textbf{3.45} & & 61.5 & 43.2 & 52.3 \\
\rowcolor{ourgray}
ReKV + FrameBlend & & \textbf{62.4} & \textbf{3.99} & & \textbf{50.1} & 3.41 & & \textbf{63.0} & \textbf{44.6} & \textbf{53.8} \\
LiveVLM & & 60.1 & 3.95 & & 51.5 & 3.50 & & 63.6 & 43.5 & 53.5 \\
\rowcolor{ourgray}
LiveVLM + FrameBlend & & \textbf{62.8} & \textbf{4.01} & & \textbf{51.7} & \textbf{3.53} & & \textbf{65.8} & \textbf{43.6} & \textbf{54.7} \\
HERMES & & 60.7 & 3.97 & & 54.4 & 3.56 & & 65.7 & \textbf{48.9} & 57.3 \\
\rowcolor{ourgray}
HERMES + FrameBlend & & \textbf{62.0} & \textbf{4.00} & & \textbf{56.2} & \textbf{3.61} & & \textbf{66.6} & 48.4 & \textbf{57.5} \\
\hline
\noalign{\renewcommand{\arraystretch}{1.0}}
\multicolumn{11}{l}{\bf Qwen2.5-VL-7B} \\
\noalign{\renewcommand{\arraystretch}{1.5}}
\hline
ReKV & & 66.6 & 4.02 & & 60.1 & 3.63 & & 62.5 & \textbf{45.3} & 53.9 \\
\rowcolor{ourgray}
ReKV + FrameBlend & & \textbf{69.7} & \textbf{4.05} & & \textbf{62.6} & \textbf{3.67} & & \textbf{65.4} & 44.3 & \textbf{54.9} \\
HERMES & & 68.0 & \textbf{4.08} & & 61.7 & 3.65 & & 66.7 & 48.2 & 57.4 \\
\rowcolor{ourgray}
HERMES + FrameBlend & & \textbf{71.2} & 4.06 & & \textbf{64.6} & \textbf{3.70} & & \textbf{67.8} & \textbf{48.2} & \textbf{58.0} \\
\noalign{\hrule height 0.6pt}
\end{tabular}
\end{center}
\end{table}

\textbf{Streaming video understanding.}
As shown in Table~\ref{tab:acc_rvs_ovo_a5000_opt}, FrameBlend improves the understanding performance of the backbone frameworks.
On LLaVA-OV-7B, it improves RVS-Ego accuracy by 1.3--3.1 pp, RVS-Movie accuracy by 0.2--1.8 pp, and OVO-Bench average accuracy by 0.2--1.5 pp across ReKV, HERMES, and LiveVLM.
Similar improvements are observed on Qwen2.5-VL-7B, showing that FrameBlend consistently improves streaming performance across different backbone video-LLMs.
% While performance slightly decreases on some OVO-Bench Backward subsets, the overall OVO-Bench average consistently improves.
These results demonstrate that FrameBlend is a simple yet effective method that can be applied to diverse SVU frameworks.

\begin{table}[t]
\caption{\textbf{FrameBlend on offline benchmarks.} We compare the accuracy of various SVU frameworks with and without FrameBlend on offline VQA benchmarks.}
\label{tab:acc_offline_a5000_opt}
\begin{center}
\small
\renewcommand{\arraystretch}{1.30}
\setlength{\tabcolsep}{2.9pt}
\begin{tabular}{l c c c c c c c c c c c c} 
\noalign{\hrule height 0.6pt}
 & & \multicolumn{4}{c}{\bf Video-MME} & & \multicolumn{4}{c}{\bf MLVU} & & {\bf EgoSchema} \\
\cline{3-6} \cline{8-11} \cline{13-13}
{\bf LLaVA-OV-7B} & & Short & Medium & Long & Total & & Holistic & {\rule[-0.6ex]{0pt}{4.0ex}\renewcommand{\arraystretch}{1.0}\begin{tabular}[c]{@{}c@{}}Single-\\Detail\end{tabular}} & {\renewcommand{\arraystretch}{1.0}\begin{tabular}[c]{@{}c@{}}Multi-\\Detail\end{tabular}} & Total & & Total \\
\hline
ReKV & & \textbf{71.2} & 55.3 & 46.9 & 57.8 & & 83.2 & 73.1 & 41.1 & 68.4 & & 60.0 \\
\rowcolor{ourgray}
ReKV + FrameBlend & & 70.7 & \textbf{57.0} & \textbf{50.7} & \textbf{59.4} & & \textbf{83.6} & \textbf{73.4} & \textbf{41.3} & \textbf{68.7} & & \textbf{60.3} \\
LiveVLM & & 68.4 & \textbf{56.2} & 48.4 & 57.7 & & 82.7 & 70.8 & 41.1 & 67.0 & & \textbf{60.6} \\
\rowcolor{ourgray}
LiveVLM + FrameBlend & & \textbf{69.6} & 56.1 & \textbf{48.8} & \textbf{58.1} & & \textbf{83.4} & \textbf{71.9} & \textbf{42.8} & \textbf{68.1} & & 60.5 \\
HERMES & & 70.4 & 52.0 & 44.9 & 55.8 & & 81.2 & 61.5 & 35.7 & 60.2 & & 59.7 \\
\rowcolor{ourgray}
HERMES + FrameBlend & & \textbf{71.1} & \textbf{57.4} & \textbf{47.6} & \textbf{58.7} & & \textbf{83.4} & \textbf{65.4} & \textbf{41.3} & \textbf{64.1} & & \textbf{61.3} \\
\noalign{\hrule height 0.6pt}
\end{tabular}
\end{center}
\end{table}

\textbf{Offline video understanding.}
Following prior work~\citep{di2025streaming,zhang2026hermes}, we further evaluate FrameBlend on offline benchmarks to demonstrate its general video understanding capability.
As shown in Table~\ref{tab:acc_offline_a5000_opt}, FrameBlend improves accuracy by 0.3--3.9 pp on MLVU and 0.4--2.9 pp on Video-MME across ReKV, HERMES, and LiveVLM.
On EgoSchema, it maintains comparable performance or improves it, with changes ranging from $-0.1$ to $+1.6$ pp.
Overall, these results demonstrate that FrameBlend improves the general video understanding capability of SVU frameworks beyond streaming settings.

\subsection{Efficiency Analysis} \label{ssec:efficiency}

\begin{table}[t]
\caption{\textbf{Efficiency metrics.} We compare SVU methods with and without FrameBlend on RVS-Ego.}
\label{tab:eff_ego_latency_a100_pool}
\begin{center}
\small
\renewcommand{\arraystretch}{1.30}
\setlength{\tabcolsep}{4.45pt}
\begin{tabular}{l c cc c cc c cc} 
\noalign{\hrule height 0.6pt}
 & & \multicolumn{2}{c}{\bf Prefill (idle time)} & & \multicolumn{2}{c}{\bf Answering (query time)} & & \multicolumn{2}{c}{\bf Total} \\
\cline{3-4} \cline{6-7} \cline{9-10}
{\bf LLaVA-OV-7B} & & {TFLOPs/f} & { ms/f} & & { Recomp. (s)} & { TTFT (s)} & & { TFLOPs} & { Latency (s)} \\
\hline
ReKV & & 4.38 & 81.1 & & -- & 0.17 & & 4647.8 & 90.6 \\
\rowcolor{ourgray}
ReKV + FrameBlend & & \textbf{3.24} & \textbf{32.0} & & 0.77 & 0.93 & & \textbf{3567.4} & \textbf{37.4} \\
% \hline
% \hline
LiveVLM & & 3.78 & 32.6 & & -- & 0.04 & & 4007.2 & 37.6 \\
\rowcolor{ourgray}
LiveVLM + FrameBlend & & \textbf{2.94} & \textbf{28.5} & & 0.77 & 0.84 & & \textbf{3247.6} & \textbf{32.3} \\
HERMES & & 4.07 & 46.4 & & -- & 0.03 & & 4333.0 & 51.7 \\
\rowcolor{ourgray}
HERMES + FrameBlend & & \textbf{3.57} & \textbf{37.7} & & 0.95 & 1.04 & & \textbf{3914.3} & \textbf{42.0} \\
\noalign{\hrule height 0.6pt}
\end{tabular}
\end{center}

\vspace{-2.5mm}

\end{table}

\begin{figure}[t]
\centering
\begin{subfigure}[t]{0.49\linewidth}
    \centering
    \includegraphics[width=\linewidth]{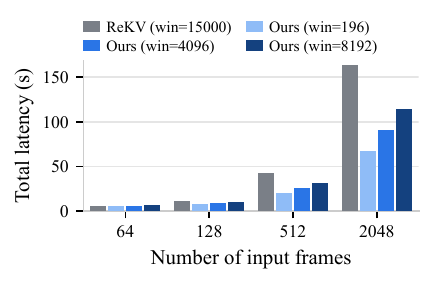}
    \caption{Total latency}
    \label{fig:latency_by_frame}
\end{subfigure}
\hfill
\begin{subfigure}[t]{0.49\linewidth}
    \centering
    \includegraphics[width=\linewidth]{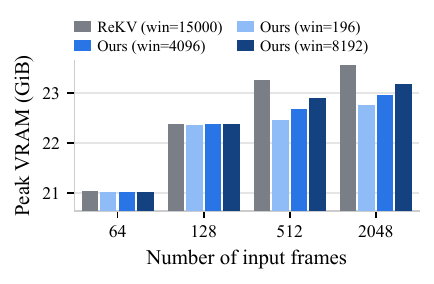}
    \caption{Peak VRAM}
    \label{fig:vram_by_frame}
\end{subfigure}
\caption{\textbf{Efficiency at varying numbers of input frames.} We analyze ReKV with FrameBlend under varying video lengths, and observe that the efficiency gain increases as the video gets longer.
}
\label{fig:latency_vram_by_frame}

\vspace{-1mm}

\end{figure}

\begin{figure}[t]
\centering
\begin{minipage}[t]{0.32\linewidth}
    \centering
    \includegraphics[width=\linewidth]{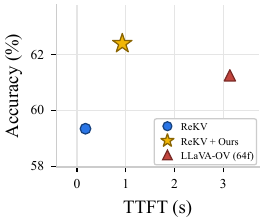}\\
    {\small (a) RVS-Ego}
\end{minipage}%
\hfill
\begin{minipage}[t]{0.32\linewidth}
    \centering
    \includegraphics[width=\linewidth]{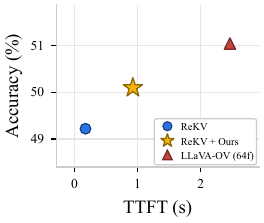}\\
    {\small (b) RVS-Movie}
\end{minipage}
\hfill
\begin{minipage}[t]{0.32\linewidth}
    \centering
    \includegraphics[width=\linewidth]{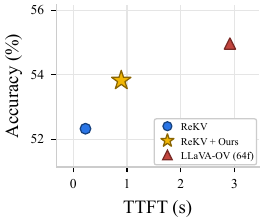}\\
    {\small (c) OVO-Bench}
\end{minipage}%
\caption{\textbf{TTFT--accuracy trade-off.} We compare ReKV with and without FrameBlend, and the offline LLaVA-OV-7B using 64 uniformly sampled frames. 
}
\label{fig:ttft_acc_rekv}
\end{figure}

 % comparing with llava-ov-7b (64 frames)

\textbf{Idle- and query-time efficiency.}
We evaluate the efficiency of FrameBlend on RVS-Ego by separately measuring idle-time and query-time costs.
For idle-time efficiency, we report cost on a per-frame basis, as SVU continuously processes incoming frames before a query arrives.
We analytically estimate FLOPs and measure wall-clock latency, reported as TFLOPs per frame and milliseconds per frame, respectively.
For query-time efficiency, we report recomputation latency and time to first token (TTFT) per question.
Since RVS-Ego contains multiple questions in each video, we evaluate each question independently using all video frames preceding that question as input.

As shown in Table~\ref{tab:eff_ego_latency_a100_pool}, FrameBlend reduces per-frame prefill FLOPs by 12--26\% and prefill latency by 13--61\% across ReKV, HERMES, and LiveVLM.
Although recomputation increases TTFT to approximately one second, the savings in idle-time computation exceed the additional query-time cost: total FLOPs decrease by 10--23\% and total latency by 14--59\%.
These savings mainly result from the shorter prefill window enabled by recomputation, which reduces the attention computation required for each incoming frame.

\textbf{Efficiency across different video lengths.}
Since the benefit of reducing per-frame prefill cost accumulates over a video stream, we further evaluate efficiency as the number of input frames increases.
Figure~\ref{fig:latency_by_frame} shows that FrameBlend increasingly reduces total latency over ReKV as the video grows longer, due to its shorter prefill window.
Similarly, peak VRAM in Figure~\ref{fig:vram_by_frame} exhibits greater memory savings as the video grows longer.
This is because, in ReKV, the shorter prefill window keeps fewer KVs resident on the GPU while offloading older KVs to host memory.
Additionally, peak VRAM is similar between ReKV and FrameBlend for short inputs, where memory usage is dominated by vision encoder activations.

\textbf{Comparison with query-time prefill.}
We also compare FrameBlend with an offline LLaVA-OV-7B baseline that uniformly samples and prefills 64 frames at query time.
This places FrameBlend between two settings: ReKV minimizes TTFT by directly reusing cached KVs, whereas the offline baseline performs the full prefill after the query arrives.
As shown in Figure~\ref{fig:ttft_acc_rekv}, FrameBlend incurs additional TTFT over ReKV due to recomputation, but remains substantially faster than the offline baseline while achieving higher accuracy than ReKV.
Thus, FrameBlend provides a balanced TTFT--accuracy trade-off by introducing selective recomputation to SVU frameworks.

\subsection{Recomputation target selection analysis} \label{ssec:analysis}

\begin{table}[t]
\caption{\textbf{Comparison against CacheBlend.}
Accuracy and TTFT when applied on ReKV. 
}
\label{tab:acc_ttft_cacheblend}
\begin{center}
\small
\renewcommand{\arraystretch}{1.30}
\setlength{\tabcolsep}{4.25pt}
\begin{tabular}{l c cc c cc c cc}  
\noalign{\hrule height 0.6pt}
 & & \multicolumn{2}{c}{\bf OVO-Bench} & & \multicolumn{2}{c}{\bf RVS-Movie} & & \multicolumn{2}{c}{\bf RVS-Ego} \\
\cline{3-4} \cline{6-7} \cline{9-10}
{\bf LLaVA-OV-7B} & & {Acc.} & {Added TTFT (s)} & & {Acc.} & {Added TTFT (s)} & & {Acc.} & {Added TTFT (s)} \\
\hline
ReKV & & 52.3 & -- & & 49.2 & -- & & 59.3 & -- \\
ReKV + CacheBlend & & 52.7 & 0.90 & & 48.0 & 1.51 & & 61.4 & 2.26 \\
\rowcolor{ourgray}
ReKV + FrameBlend & & \textbf{53.8} & \textbf{0.67} & & \textbf{50.1} & \textbf{0.75} & & \textbf{62.4} & \textbf{0.76} \\
\noalign{\hrule height 0.6pt}
\end{tabular}
\end{center}
\end{table}

We compare QTS with CacheBlend~\citep{yao2025cacheblend} to evaluate the effect of recomputation target selection.
For a controlled comparison, we follow the released CacheBlend implementation while using the same prefill window size and KV replacement policy as FrameBlend. This setup isolates the effect of target selection: the CacheBlend-based variant fully recomputes early layers to identify targets in a query-agnostic manner, whereas QTS directly selects query-relevant targets without such additional computation.
As shown in Table~\ref{tab:acc_ttft_cacheblend}, FrameBlend consistently achieves higher accuracy with lower added TTFT across all streaming benchmarks. These results show that QTS provides a more favorable accuracy--latency trade-off for selective recomputation in SVU.

\subsection{Ablation Study} \label{ssec:ablation}

\begin{table}[!t]
\caption{\textbf{Ablation studies.} We ablate two key components of FrameBlend.}
\label{tab:acc_ablation}
\begin{center}
\small
\renewcommand{\arraystretch}{1.30}
\setlength{\tabcolsep}{7.35pt}
\begin{tabular}{l c cc c cc c ccc} 
\noalign{\hrule height 0.6pt}
 & & \multicolumn{2}{c}{\bf RVS-Ego} & & \multicolumn{2}{c}{\bf RVS-Movie} & & \multicolumn{3}{c}{\bf OVO-Bench} \\
\cline{3-4} \cline{6-7} \cline{9-11}
{\bf LLaVA-OV-7B} & & Acc. & Score & & Acc. & Score & & Real-time & Backward & Avg. \\
\hline
ReKV & & 59.3 & 3.93 & & 49.2 & 3.45 & & 61.5 & 43.2 & 52.3 \\
\quad + short window & & 60.7 & 3.94 & & 46.8 & 3.30 & & \textbf{63.3} & 43.0 & 53.1 \\
\quad + recomputation & & 58.7 & 3.94 & & 48.9 & \textbf{3.46} & & 61.7 & \textbf{44.8} & 53.2 \\
\rowcolor{ourgray}
\quad + both & & \textbf{62.4} & \textbf{3.99} & & \textbf{50.1} & 3.41 & & 63.0 & 44.6 & \textbf{53.8} \\
\noalign{\hrule height 0.6pt}
\end{tabular}
\end{center}
\end{table}

We ablate the two key components of FrameBlend on ReKV: a shorter attention window during idle-time prefill and selective recomputation at query time.
As shown in Table~\ref{tab:acc_ablation}, applying either component alone has mixed effects across benchmarks.
A shorter prefill window removes part of the inter-token context, which can reduce unnecessary context but may also discard useful context for answering the query.
Query-time recomputation complements this by selectively restoring query-relevant context among the selected tokens.
Consequently, combining the shorter prefill window with recomputation achieves the highest overall accuracy across all benchmarks.

\section{Conclusion}

In this work, we introduced FrameBlend, a simple yet effective plug-and-play method for SVU.
FrameBlend addresses the context mismatch in prior SVU frameworks by selectively recomputing cached KVs at query time without requiring any training.
To enable efficient recomputation under the tight query-time latency budget of SVU, we propose QTS, which selects query-relevant recomputation targets with minimal additional overhead.
We further show that, with query-time recomputation, using shorter attention windows during idle-time prefill can improve performance while substantially reducing computation.
Evaluated across three SVU frameworks and five streaming and offline video understanding benchmarks, FrameBlend consistently improves (or matches) performance while reducing the overall computational cost.

\textbf{Limitations.}
While FrameBlend improves accuracy and overall efficiency, several limitations remain and could be addressed in future work.
Although we evaluate FrameBlend across multiple SVU frameworks, video-LLM backbones, and benchmarks, our experiments are limited to 7B-scale backbone models, and its effectiveness on both smaller and larger models remains to be explored.
In addition, FrameBlend currently uses a fixed prefill window selected for each framework and benchmark; automatically adapting the window to different video streams could further improve its practicality.
Finally, although FrameBlend reduces overall computation, selective recomputation introduces additional query-time latency. Dynamically adjusting the recomputation budget according to latency constraints could further improve its applicability to settings with stricter query-time latency requirements.

\subsection*{AI use statement}

In this work, we used generative AI tools to help develop theoretical models or conceptual frameworks, propose or refine hypotheses, design or provide feedback on research methodology or experiments, implement methods, assist with translation, and interpret experimental results.
We have not used generative AI tools to generate synthetic datasets, formulate mathematical claims, provide critical ingredients for proving mathematical claims, assist in writing proofs, clean or reformat datasets, or support qualitative and thematic data analysis.
Additionally, we used generative AI tools to create or modify scientific figures or images, create or edit software code, create research artifacts, draft parts of the paper, summarize or analyze existing literature, discover research topics or identify gaps, brainstorm research ideas, source or search for information, improve the readability, and identify relevant literature.
We have reviewed all AI-assisted work.
We carefully reviewed AI-assisted hypotheses, interpretations of experimental results, scientific figures, code, and other research artifacts for correctness.
We take responsibility for the final content of this work, including text, claims, or artifacts produced with the aid of generative AI.

\subsection*{Reproducibility statement}

We provide framework-specific implementation details in Appendix~\ref{app:implementation_details}, prefill-window configurations in Appendix~\ref{app:window_size_tuning}. Section~\ref{sec:experiment} describes the benchmarks, backbone models, hyperparameters, and hardware used in our experiments. Together, these details specify the experimental settings and evaluation procedures used to reproduce our results.

\bibliography{references}
\bibliographystyle{iclr2027_conference}

\newpage

\appendix

\section{Framework-Specific Implementation Details} \label{app:implementation_details}

\subsection{ReKV with FrameBlend}
ReKV~\citep{di2025streaming} stores the KVs of all incoming frames and independently retrieves the top 64 frames at each LLM layer when a query arrives.
During idle-time prefill, the original ReKV lets each token attend to up to 15,000 preceding video tokens using a sliding-window attention.
We apply the short prefill window of FrameBlend by changing only this window size; the retrieval and decoding procedures remain unchanged.

For recomputation target selection, we directly use the layer-wise retrieval results produced by ReKV.
Because ReKV retrieves complete frame blocks, all tokens within a frame block share the same retrieval score; QTS therefore operates at the frame level.
We first rank each frame by the number of layers in which it is retrieved.
Ties are broken by the sum of its normalized retrieval scores over the layers that retrieve it.
The selected frames are ordered temporally and jointly re-prefilled through all LLM layers.
At each layer, only frames that are both selected for recomputation and present in that layer's original retrieval set are replaced by their recomputed KVs.
Therefore, FrameBlend does not add frames to the layer-wise decoding context or modify ReKV's retrieval policy.

\subsection{LiveVLM with FrameBlend}
LiveVLM~\citep{ning2025livevlm} processes incoming frames while maintaining a compressed KV cache.
It applies compression when the cache exceeds its budget of 12,000 tokens.
At query time, it independently retrieves 40\% of 16-token pages for each layer and head.
The resulting query-time context can therefore contain different video tokens across both layers and heads.

FrameBlend uses sliding-window attention with a separate, small KV buffer, allowing each incoming token to attend only to the most recent tokens of the original stream within the prefill window.
The separate KV buffer ensures that the effective window size does not depend on how much the KV cache has been compressed.
For QTS, we use the tokens selected by LiveVLM's retrieval mechanism as candidates.
Because LiveVLM has different retrieval sets for each layer--head pair, we count how many layer--head pairs retrieve each token.
We then select the most frequently retrieved tokens based on this count, using normalized retrieval scores for tie-breaking.
The selected tokens are jointly re-prefilled, and only their corresponding slots in the temporary query-time cache are replaced.

\subsection{HERMES with FrameBlend}
HERMES~\citep{zhang2026hermes} maintains a compressed KV cache instead of performing query-time retrieval.
We set the maximum KV cache size to 6,000 video tokens, and HERMES independently compresses the cache at each layer to this budget.
The original HERMES allows incoming tokens to attend to the entire compressed cache.
For FrameBlend, as in LiveVLM, we restrict this prefill attention to the most recent video tokens within the prefill window using sliding-window attention.
We also maintain a separate KV buffer containing the uncompressed KVs of these recent tokens for the same reason as in LiveVLM.

Because HERMES has no query-time retrieval, we perform an auxiliary retrieval forward pass only for recomputation target selection.
At each layer, we retrieve the top-$M$ video tokens using their query similarity and use these layer-wise sets as the voting sets for QTS.
We use $M=2{,}048$ for RVS and $M=4{,}096$ for OVO-Bench and the offline benchmarks.
We then rank tokens by the number of layers that select them and choose the most frequently retrieved tokens.
This auxiliary retrieval affects only which KVs are recomputed; the decoding context remains the full compressed KV cache produced by HERMES.
The selected tokens are re-prefilled in temporal order using the layer-specific position indices maintained by HERMES, and the recomputed KVs replace only the corresponding surviving slots in each layer.

\section{Prefill window size tuning} \label{app:window_size_tuning}

We tune the size of the short attention window for idle-time prefill separately for each framework and benchmark.
The original attention ranges are 15,000 tokens for ReKV, 12,000 tokens for LiveVLM, and 6,000 tokens for HERMES in our experiments.
Table~\ref{tab:prefill_window_sizes} summarizes the window sizes used for the main and appendix results.
Across the evaluated settings, FrameBlend uses only 1.3--54.6\% of ReKV's original attention range, 1.6--34.1\% of LiveVLM's, and 8.5--68.3\% of HERMES's.

\begin{table*}[t]
\caption{\textbf{Prefill window sizes used by FrameBlend.} Values denote the number of video tokens.}
\label{tab:prefill_window_sizes}
\begin{center}
\small
\renewcommand{\arraystretch}{1.30}
\setlength{\tabcolsep}{5.0pt}
\begin{tabular}{l c c c c c c c}
\noalign{\hrule height 0.6pt}
& \multicolumn{3}{c}{\bf Streaming benchmarks}
& & \multicolumn{3}{c}{\bf Offline benchmarks} \\
\cline{2-4} \cline{6-8}
{\bf Method}
& OVO-Bench
& RVS-Ego
& RVS-Movie
&
& Video-MME
& MLVU
& EgoSchema \\
\hline

\multicolumn{8}{l}{\bf LLaVA-OV-7B} \\
\hline
ReKV
& 2048 & 196 & 196
& & 8192 & 8192 & 384 \\

HERMES
& 1024 & 1024 & 512
& & 512 & 512 & 1024 \\

LiveVLM
& 196 & 1024 & 4096
& & 2048 & 4096 & 512 \\

\hline
\multicolumn{8}{l}{\bf Qwen2.5-VL-7B} \\
\hline
ReKV
& 4096 & 4096 & 1024
& & -- & -- & -- \\

HERMES
& 4096 & 512 & 512
& & -- & -- & -- \\

\noalign{\hrule height 0.6pt}
\end{tabular}
\end{center}
\end{table*}

\section{Additional Prefill Window Size Analysis} \label{app:window_size_sweep}

\begin{figure}[t]
\centering

% Row 1
\begin{subfigure}[t]{0.32\linewidth}
    \centering
    \includegraphics[width=\linewidth]{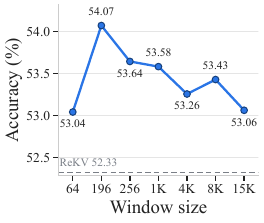}
    \caption{OVO-Bench}
    \label{fig:window_sweep_ovo}
\end{subfigure}
\hfill
\begin{subfigure}[t]{0.32\linewidth}
    \centering
    \includegraphics[width=\linewidth]{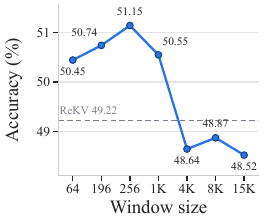}
    \caption{RVS-Movie}
    \label{fig:window_sweep_rvs_movie}
\end{subfigure}
\hfill
\begin{subfigure}[t]{0.32\linewidth}
    \centering
    \includegraphics[width=\linewidth]{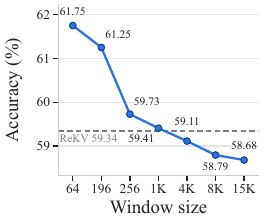}
    \caption{RVS-Ego}
    \label{fig:window_sweep_rvs_ego}
\end{subfigure}

\vspace{2mm}

% Row 2
\begin{subfigure}[t]{0.32\linewidth}
    \centering
    \includegraphics[width=\linewidth]{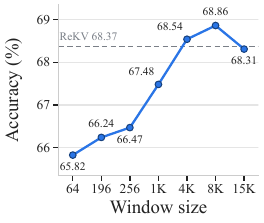}
    \caption{MLVU}
    \label{fig:window_sweep_mlvu}
\end{subfigure}
\hfill
\begin{subfigure}[t]{0.32\linewidth}
    \centering
    \includegraphics[width=\linewidth]{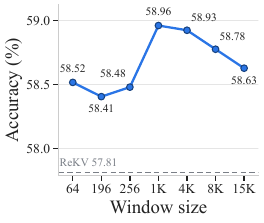}
    \caption{Video-MME}
    \label{fig:window_sweep_videomme}
\end{subfigure}
\hfill
\begin{subfigure}[t]{0.32\linewidth}
    \centering
    \includegraphics[width=\linewidth]{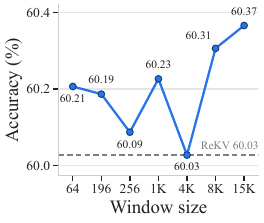}
    \caption{EgoSchema}
    \label{fig:window_sweep_egoschema}
\end{subfigure}

\caption{\textbf{Prefill window size sweep with full recomputation.}
We evaluate the accuracy of ReKV with FrameBlend across different prefill window sizes under full recomputation.}
\label{fig:full_recompute_window_sweep}
\end{figure}

To further examine how the prefill window affects performance, we sweep the window size under full recomputation ($K=\infty$) for ReKV with FrameBlend for simplicity.
As shown in Figure~\ref{fig:full_recompute_window_sweep}, performance typically peaks at a particular window size rather than improving monotonically as the window grows.
Overall, prefill windows shorter than ReKV's default window of 15K tokens tend to perform better, while the optimal window size varies across benchmarks. Even when the window is relatively large, such as 8K tokens, it is still only about half of the default window size. This also suggests that the appropriate amount of idle-time context is benchmark-dependent, making automatic adaptation of the window size to the video stream an interesting direction for future work.

\section{Additional efficiency analysis} \label{app:eff_results}

\begin{figure}[t]
\centering

% LiveVLM
\begin{subfigure}[t]{0.49\linewidth}
    \centering
    \includegraphics[width=\linewidth]{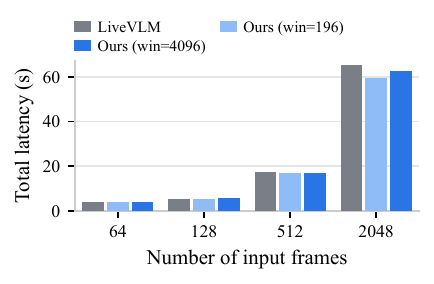}
    \caption{LiveVLM: Total latency}
    \label{fig:app_latency_by_frame_livevlm}
\end{subfigure}
\hfill
\begin{subfigure}[t]{0.49\linewidth}
    \centering
    \includegraphics[width=\linewidth]{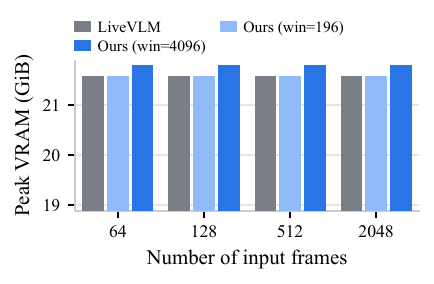}
    \caption{LiveVLM: Peak VRAM}
    \label{fig:app_vram_by_frame_livevlm}
\end{subfigure}

\vspace{2mm}

% HERMES
\begin{subfigure}[t]{0.49\linewidth}
    \centering
    \includegraphics[width=\linewidth]{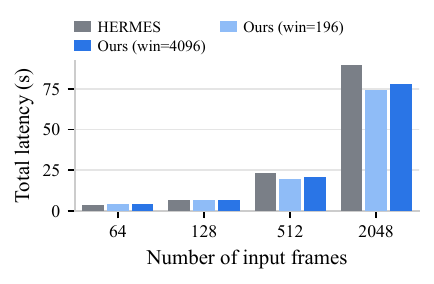}
    \caption{HERMES: Total latency}
    \label{fig:app_latency_by_frame_hermes}
\end{subfigure}
\hfill
\begin{subfigure}[t]{0.49\linewidth}
    \centering
    \includegraphics[width=\linewidth]{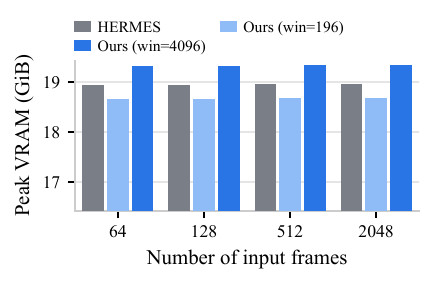}
    \caption{HERMES: Peak VRAM}
    \label{fig:app_vram_by_frame_hermes}
\end{subfigure}

\caption{\textbf{Efficiency at varying numbers of input frames.}
We analyze LiveVLM and HERMES with FrameBlend under varying video lengths
and observe that the efficiency gain increases as the video becomes longer.}
\label{fig:app_latency_vram_by_frame}
\end{figure}

Here, we provide additional efficiency results for HERMES and LiveVLM as the video stream grows.
As in the main experiments, wall-clock latency is measured on a single NVIDIA A100 80GB GPU.
As shown in Figure~\ref{fig:app_latency_vram_by_frame}, FrameBlend reduces latency for both frameworks, although the reduction is smaller than for ReKV.
This is because HERMES and LiveVLM already use compressed KV caches and therefore spend less time on attention.

For peak GPU memory, FrameBlend with a window size of 196 uses comparable or lower memory than the corresponding baseline.
With a window size of 4096, the additional KV buffer used by compression-based frameworks slightly increases peak memory, by approximately 0.38\,GiB for HERMES and 0.22\,GiB for LiveVLM.

\section{Additional CPU memory usage}
FrameBlend additionally stores projected visual embeddings in CPU memory for query-time recomputation.
On RVS-Ego, whose videos are approximately one hour long, these embeddings require 2.36\,GiB for ReKV, 0.25\,GiB for HERMES, and 1.01\,GiB for LiveVLM.
For compression-based frameworks, only embeddings corresponding to tokens that remain in at least one layer can become recomputation candidates, resulting in lower CPU memory overhead.

\end{document}